\documentclass[runningheads]{llncs}
\usepackage[T1]{fontenc}
\usepackage{amsmath}
\usepackage{graphicx}
\usepackage{xcolor}
\usepackage{subcaption}
\usepackage{hyperref}
\usepackage{float}
\usepackage{fancyhdr}

\usepackage{mdframed} 
\begin{document}

\title{NeuroPredict: EEG Based Deep Learning Framework for SSRI Response Prediction in Major Depressive Disorder}

\titlerunning{NeuroPredict}

\authorrunning{M. F. K. Chowdhury et al.}


\author{
Md Fahimul Kabir Chowdhury\inst{1} \and
Jayed Mohammad Barek\inst{1} \and
Md Ariful Hasan\inst{1} \and
Haili Wang\inst{1}
}

\institute{
Department of Computer Science, University of North Texas, Denton, TX, USA\\
\email{MdFahimulKabirChowdhury@my.unt.edu}
}

\maketitle              

\footnotetext{Published in the proceedings of
ICONIP 2026, Springer Nature.
DOI: \href{https://doi.org/10.1007/978-981-92-8146-6_11}
{10.1007/978-981-92-8146-6\_11}.}

\begin{abstract}
Major Depressive Disorder (MDD) is a leading cause of disability worldwide, yet selective serotonin reuptake inhibitors (SSRIs) yield highly variable outcomes across patients. To address the need for individualized treatment prediction, we introduce NeuroPredict, a non‑invasive EEG‑based framework that leverages rich time-frequency representations of pre‑treatment signals. EEG data were rigorously preprocessed with artifact removal and transformed using Wigner-Ville distribution (WVD), continuous wavelet transform (CWT), and constant‑Q transform (CQT), then classified with a tailored convolutional neural network (CNN). We systematically benchmarked NeuroPredict against state‑of‑the‑art pretrained models (Xception, DenseNet201, MobileNetV2) and EEG‑specific baselines (EEGNet, DeepSleepNet, DeepConvNet). Our proposed CNN consistently outperformed these approaches, achieving a peak accuracy of 90.23\% (95\% CI for AUC: 96.3-97.1\%), alongside strong sensitivity, specificity, and F1 score. 
These findings establish NeuroPredict as a scalable and clinically viable tool for precision prediction of SSRI treatment response, advancing the integration of data‑driven methods into personalized psychiatry.

\end{abstract}
\section{Introduction}
MDD is a debilitating illness that disrupts emotional well-being, cognitive processes, and everyday functioning. Selective Serotonin Reuptake Inhibitors (SSRIs) serve as the primary pharmacological intervention for this condition \cite{chowdhury2026multifrequency,stahl2021stahl}. 
While extensive meta-analyses confirm their efficacy and favorable safety profile compared to tricyclic antidepressants \cite{cipriani2018comparative}, clinical response remains highly variable across individuals. Consequently, only a subset of patients achieve full remission, with many requiring multiple treatment attempts before identifying an effective option \cite{zhou2021machine}. This trial-and-error process is time-consuming, burdensome, and places significant emotional strain on both patients and their families.

Recent work has focused on predicting antidepressant outcomes at baseline or early in treatment using machine learning, as early symptom changes can predict SSRI remission \cite{zhou2021machine,li2025neurophysiological,korani2026fusion}. Complementary studies have leveraged EEG features such as spectral power, connectivity, and entropy to capture neurophysiological markers of treatment response \cite{li2025neurophysiological}. A systematic review reports that models integrating clinical, EEG, and other biological markers outperform traditional approaches, while emphasizing the need for larger, clinically deployable frameworks \cite{yadav2025evaluating}.



In \cite{li2025neurophysiological}, an EEG‑based machine learning framework was developed using data from 27 patients with depression and validated on an independent cohort of 5. A SVM classifier achieving 96.8\% accuracy on 12‑second EEG windows. In \cite{yadav2025evaluating}, a systematic review of 30 studies examined ML approaches for predicting antidepressant response. Across diverse datasets, SVM consistently achieved reliable performance (AUC 0.65-0.74) using clinical and symptom features, while EEG‑based models reported accuracies up to 88\%.

In \cite{albizu2024machine}, individualized neuroimaging-derived computational models were applied to data from 16 ELECT trial participants, achieving over 90\% accuracy in classifying treatment responders (AUC 0.90, F1 0.92). In \cite{lin2025integrating}, a multi-polygenic score framework tested in two Taiwanese cohorts (N=177, N=245) showed modest accuracy for SSRI response prediction (AUC 0.631), which improved substantially when early symptom improvement was included (AUC 0.859). In \cite{byun2025automated}, machine learning models evaluated in 147 participants achieved moderate stress-detection accuracy (0.67–0.73), increasing markedly with longitudinal scaling (up to 0.94 for MDD and 0.96 for HCs).

In \cite{wang2025predicting}, EEG functional connectivity from 30 untreated depression patients across resting and video-watching conditions was used to predict SSRI response, achieving over 95\% accuracy in cross-validation and over 85\% on an independent test set using SVMs. In \cite{moncy2024prediction}, baseline EEG from 21 participants in a double-blinded home-based tDCS trial was analyzed with deep learning, where a 1D-CNN achieved 85.7\% accuracy (71.4\% specificity, 92.8\% sensitivity). In \cite{shahabi2025deep}, a Vision Transformer applied to pre-treatment EEG from 30 MDD patients achieved 92\% validation and 91\% test accuracy, demonstrating strong generalizability in predicting SSRI response.

Despite these advancements, a critical research gap remains in extracting meaningful features from the non-stationary dynamics of EEG signals. Prior frameworks, such as \cite{zhou2021machine} and \cite{li2025neurophysiological}, rely on static sociodemographic variables or polygenic scores, which fail to capture real-time brain states modulated by antidepressant therapy. While some deep learning approaches analyze raw 1D EEG signals, they often fail to isolate transient patterns from noise without manual feature engineering, and few report subject-wise splitting. NeuroPredict addresses these limitations by reframing EEG classification as a visual texture recognition task. By converting 1D signals into high-resolution CWT-based time–frequency images, the framework leverages CNNs to automatically identify complex oscillatory biomarkers that conventional spectral features may miss. Furthermore, recent work has demonstrated the effectiveness of computationally efficient CNN architectures for medical image classification, achieving high predictive performance while maintaining relatively low computational complexity \cite{chowdhury2025brain}. NeuroPredict builds on this principle by employing an efficient CNN architecture tailored to EEG-derived time–frequency representations and evaluating performance with subject-wise splitting to ensure fair assessment.


The main contributions of this work are as follows:

\begin{itemize}

\item We introduce NeuroPredict, a non‑invasive EEG‑based framework for forecasting SSRI treatment response in MDD, leveraging rich time–frequency representations rather than relying solely on clinical or imaging variables.

\item We investigate multiple time–frequency transformations (WVD, CWT, CQT) applied to rigorously preprocessed EEG signals, supported by modules for acquisition, artifact removal, and representation.

\item To validate the proposed CNN, we perform comparisons against state‑of‑the‑art (Xception, DenseNet201, MobileNetV2) and EEG‑specific baselines (DeepSleepNet, EEGNet, DeepConvNet), demonstrating clear performance advantages.

\item We present a comprehensive evaluation of NeuroPredict, reporting accuracy, F1 score, Sensitivity, Specificity, and ROC metrics for SSRI outcome prediction.

\end{itemize}

\section{Methodology}

\begin{figure}[t]
\centering
  {\includegraphics[width=0.8\linewidth]{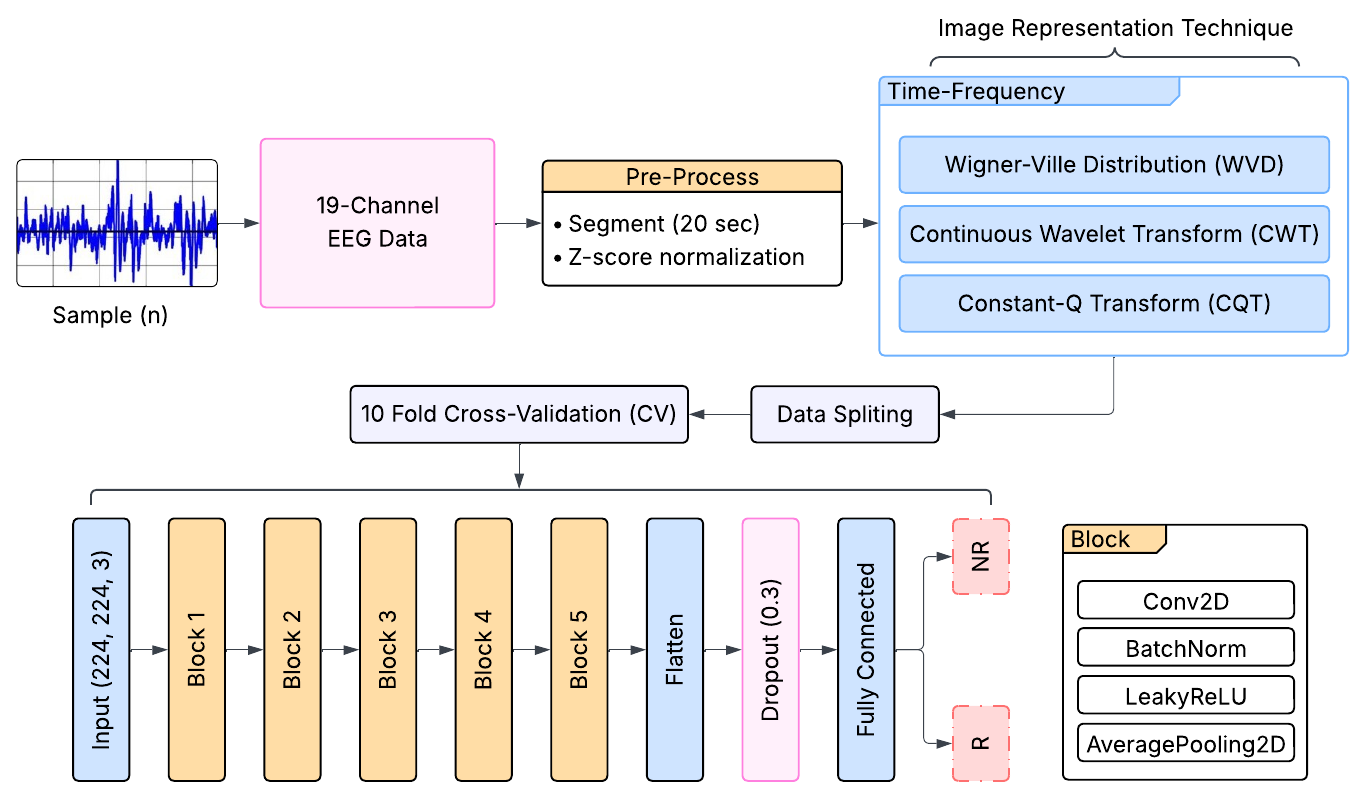}}
  \caption{Summary of EEG preprocessing and time–frequency image transformation for input to the CNN.}
  \label{fig:blockD}
\end{figure}

All experiments were conducted on a high-performance computing platform equipped with two NVIDIA H100 NVL GPUs, each offering 95,830 MB of dedicated memory and running CUDA 12.2. This computational infrastructure provided the capacity necessary for efficient model training, fine-tuning, and evaluation throughout the study. A schematic representation of the proposed framework is shown in Fig.~\ref{fig:blockD}. 

\subsection{EEG Data Acquisition}

In this study, we examined pre-treatment resting-state EEG recordings from thirty people with MDD while their eyes were closed (EC) \cite{mumtaz2017wavelet}, publicly available at \cite{Mumtaz2016}. The study protocol was approved by the Human Ethics Committee of Hospital University Sains Malaysia, Kelantan, and diagnoses followed the DSM-IV criteria. Among the participants, 12 were classified as treatment responders, defined as a $\geq 50\%$ reduction in Beck Depression Inventory (BDI) scores after antidepressant therapy relative to baseline. A 19-channel montage based on the worldwide 10–20 system with linked-ear reference was used to capture the EEG for five minutes. Power-line noise at 50 Hz was removed using a notch filter, followed by band-pass filtering between 0.5–70 Hz. Signals were digitized at 256 Hz. Responder (R) and non-responder (NR) segments were produced by dividing each 300-second recording into non-overlapping 15-second periods. The clinical features of the subjects are summarized in Table  \ref{tab:mdd_clinical_summary}.

\begin{table}[htbp]
  \centering
  \scriptsize
  \caption{Demographic and clinical overview of MDD participants grouped by treatment response.}
  \label{tab:mdd_clinical_summary}
  \begin{tabular}{|c|c|c|c|}
  \hline
    \textbf{Characteristic} & \textbf{Responder (R)} & 
    \textbf{Non-Responder (NR)} & \textbf{Total} \\
    \hline
    Age (years) & 40.7 $\pm$ 13.0 & 41.1 $\pm$ 12.5 & 40.3 $\pm$ 12.9 \\
    \hline
    Gender (Male/Female) & 8 / 8 & 9 / 9 & 17 / 17 \\
    \hline
    BDI-II (Pre-Treatment) & 18.4 $\pm$ 7.4 & 22.8 $\pm$ 12.5 & 20.6 $\pm$ 8.6 \\
    \hline
    BDI-II (Post-Treatment) & 9.1 $\pm$ 6.3 & 22.1 $\pm$ 3.3 & 15.6 $\pm$ 4.5 \\
    \hline
  \end{tabular}
\end{table}

\subsection{Time-frequency(TF) methods}

\begin{figure}[htbp]
\centering
  {%
    \begin{tabular}{ccc}
      \includegraphics[width=0.25\linewidth]{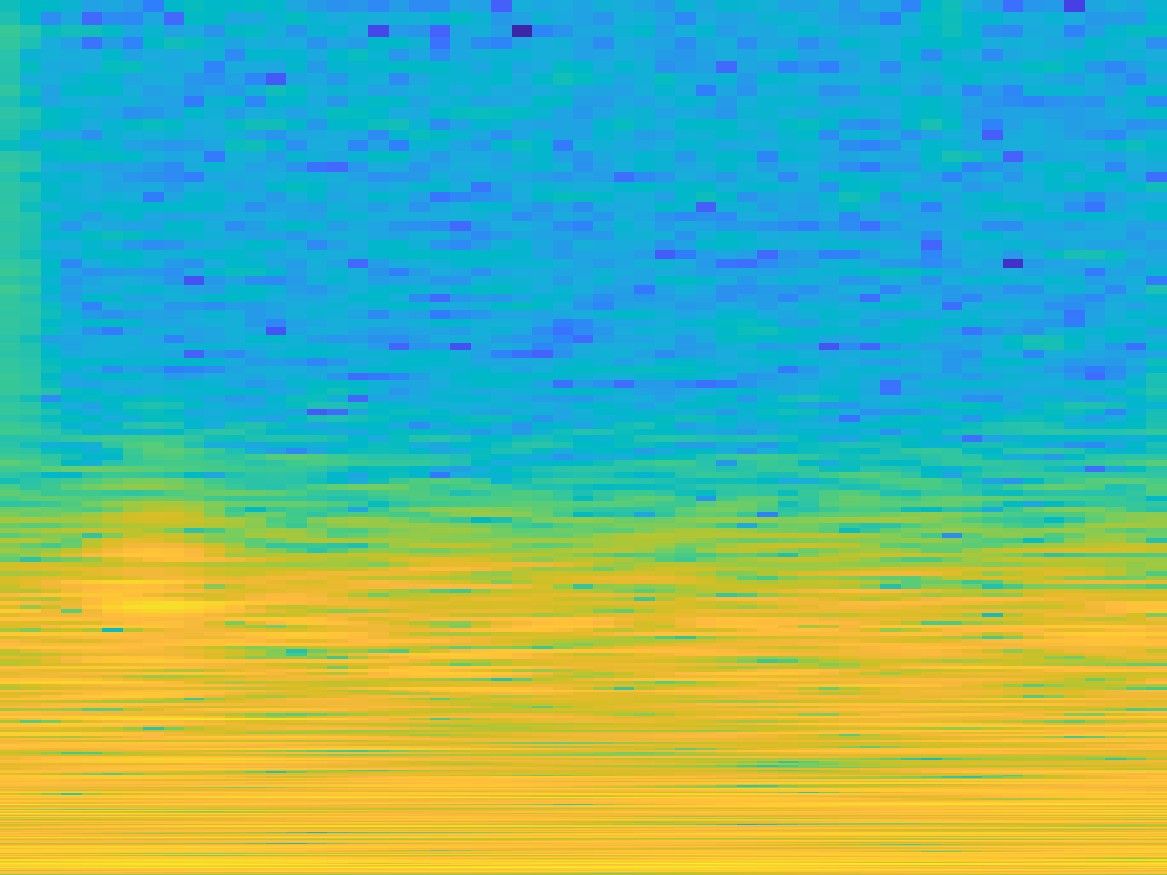} &
      \includegraphics[width=0.25\linewidth]{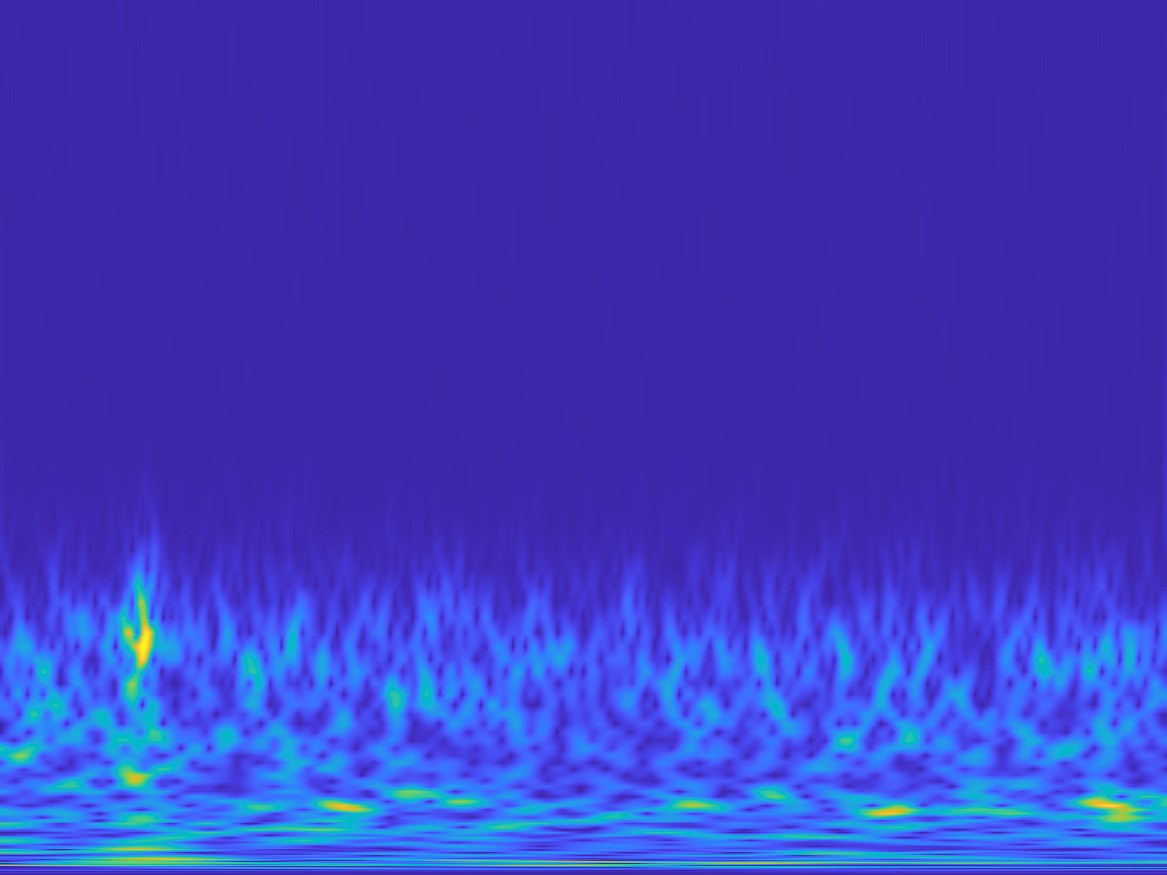} &
      \includegraphics[width=0.25\linewidth]{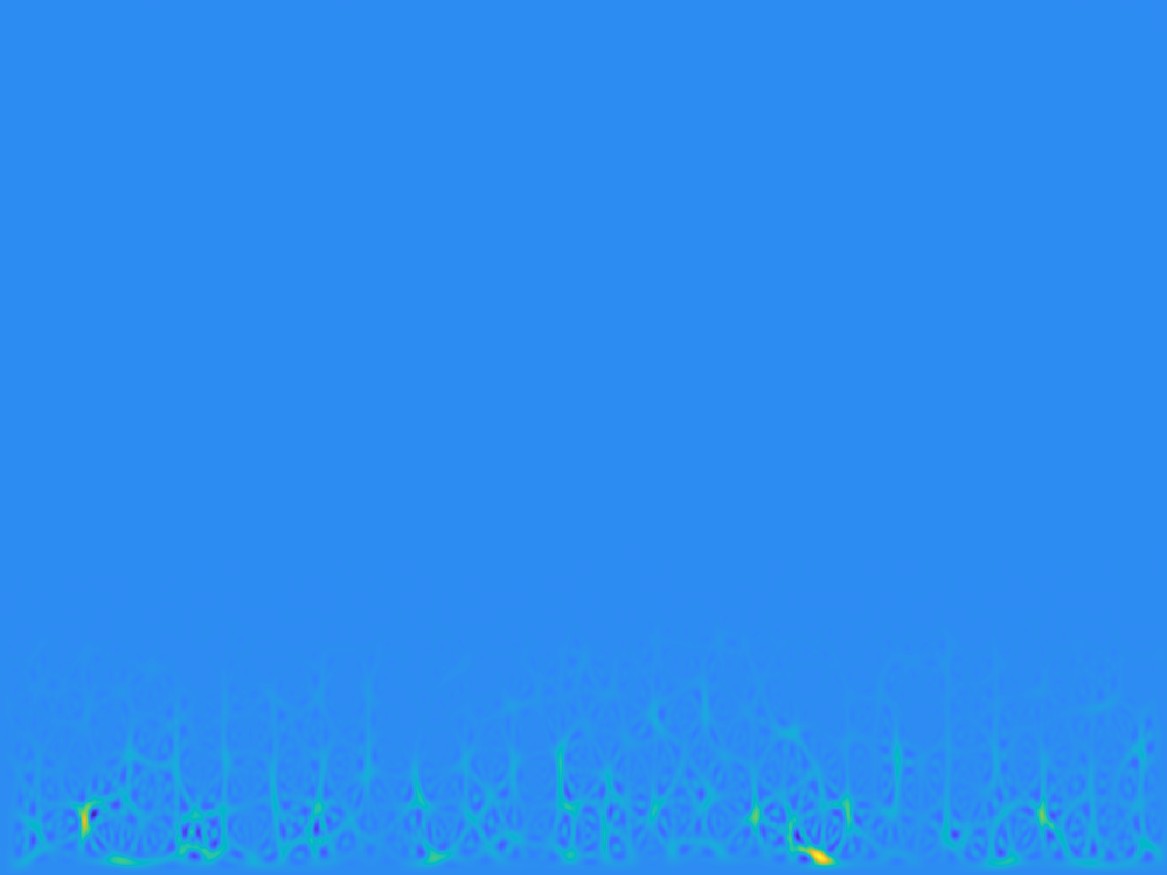} \\
      CQT & CWT & WVD
    \end{tabular}
  }
  
  \caption{Examples of time-frequency representations generated from the EEG.}
  \label{fig:tfr_examples}
\end{figure}

By converting the EEG data into the picture domain using three different time-frequency analysis techniques: CQT, CWT, and WVD-CNNs are used for EEG categorization. These methods transform the data into time-frequency representations, which are subsequently fed into the CNN seen in Fig. \ref{fig:tfr_examples}.

\subsubsection*{Constant-Q Transform (CQT)}

The CQT is a time-frequency analysis technique designed to produce frequency bins with a constant ratio between frequency and resolution. 
Unlike traditional Fourier transforms, the CQT offers a logarithmic frequency scale, making it especially suited for signals with varying frequencies. 
For EEG signals, CQT converts the one-dimensional time-domain signal $x(t)$ into a two-dimensional time-frequency representation, where the time axis corresponds to the signal’s temporal evolution and the frequency axis corresponds to the varying frequency bins. The CQT of a signal can be computed as:

\begin{equation}
\text{CQT}(t, f) = \int_{-\infty}^{\infty} x(t) \, g(t, f) \, dt
\end{equation}

where \( g(t, f) \) is the analysis window with a frequency-dependent resolution. 

\subsubsection*{Continuous Wavelet Transform (CWT)}

The CWT is another time-frequency analysis method that offers a multi-resolution representation of the signal. The CWT transforms a one-dimensional EEG signal into a two-dimensional time-frequency image by convolving the signal with a set of wavelet functions at different scales and positions. Mathematically, the CWT is defined as:

\begin{equation}
\text{CWT}(t, \tau, s) = \int_{-\infty}^{\infty} x(t') \, \psi^* \left( \frac{t' - t}{s} \right) \, dt'
\end{equation}

where \( \psi \) is the wavelet function, \( s \) is the scale parameter, and \( t \) is the time index. 
The CWT provides a detailed time-frequency representation with the flexibility to adjust the scale (or frequency) of the wavelet, making it effective for analyzing non-stationary signals like EEG.

\subsubsection*{Wigner-Ville Distribution (WVD)}

The Wigner-Ville Distribution (WVD) is a time-frequency method that provides a high-resolution representation of a signal by calculating its instantaneous frequency content. It is defined as the Fourier transform of the autocorrelation function of the signal and is given by:

\begin{equation}
\text{WVD}(t, f) = \int_{-\infty}^{\infty} x(t + \tau / 2) x^*(t - \tau / 2) e^{-i 2 \pi f \tau} \, d\tau
\end{equation}

where \( x(t) \) is the EEG signal and \( f \) is the frequency. 
The WVD provides a precise time-frequency representation with sharp frequency localization, making it well-suited for capturing instantaneous spectral changes in EEG signals. 

\subsection{Data Splitting and Cross-Validation Setup}
To ensure a robust and clinically relevant evaluation, we implemented a subject-independent 10-fold cross-validation protocol. Unlike random segment shuffling, which can lead to data leakage, to make sure that every recording from a particular patient was rigorously segregated in the training, validation, or test sets, we grouped EEG segments by participant ID.

In each fold, the dataset was partitioned such that approximately 10\% of the subjects were held out for testing, remaining 90\% were used for training. Among the training data, 15\% goes to validation (to monitor overfitting and perform early stopping), and the remaining 85\% remains for actual training. This splitting process was automated using \texttt{GroupKFold} logic from \texttt{sklearn.model\_selection}, which ensures that the model is evaluated on strictly unseen patients, thereby providing a realistic estimate of its generalization ability in a clinical setting.

\subsection{Proposed CNN Architecture}

The suggested CNN architecture, seen in Fig. \ref{fig:blockD}, is intended to categorize SSRI treatment responders and non-responders. The model uses convolutional layers with increasing filter sizes (32, 64, 128, 256, and 512) to handle input pictures of size (224 $\times$ 224 $\times$ 3). Each convolutional layer uses a kernel size of \(5 \times 5\), followed by batch normalization, LeakyReLU activation, and average pooling (\(2 \times 2\)) to reduce spatial dimensions while retaining key features.  

The final convolutional block is flattened into a 1D vector, followed by dropout (0.3) to reduce overfitting. A fully connected layer with a single sigmoid neuron performs binary classification. The model is compiled with the AdamW optimizer (learning rate \(1\times10^{-4}\)), binary cross-entropy loss, and accuracy as the evaluation metric.

\subsection{Transfer Learning (TL) Architecture}

To fairly compare with the proposed CNN, several benchmark pre-trained models were implemented, including DenseNet201, MobileNetV2, and Xception. All models used ImageNet-initialized weights and a fixed input shape of $224 \times 224 \times 3$. Each model employed a consistent classification head: LeakyReLU activation, dropout (0.5), a fully linked dense layer with 1024 units, a global average pooling layer, and a sigmoid output layer for binary-class classification. Maintaining a uniform head across networks isolated the impact of the feature extractors, minimizing variations due to architectural differences.

\subsection{EEG-Specific Baseline Models}

Well-established EEG based deep learning architectures commonly used in EEG analysis: EEGNet, DeepConvNet, and DeepSleepNet. EEGNet~\cite{lawhern2018eegnet} offers a low computational cost while being well-suited for compact EEG representations. DeepConvNet~\cite{schirrmeister2017deep} employs a deeper sequence of convolutional layers, augmented with max-pooling and dropout, to learn robust features from multi-channel EEG data. DeepSleepNet~\cite{supratak2017deepsleepnet} leverages a combination of convolutional and recurrent layers to capture both local temporal patterns and long-range dependencies in EEG signals. This architecture enables effective modeling of sleep stage transitions across sequential epochs. 

To maintain consistency between experiments, all three models were trained using the same 10-fold cross-validation methodology and modified to analyze our time-frequency EEG inputs of size (224 $\times$ 224 $\times$ 3).

\subsection{Model evaluation and validation}
Although numerous metrics exist to evaluate the performance of classification models, accuracy remains a core and commonly adopted criterion. In this work, we focus primarily on accuracy as the principal evaluation metric:

\begin{equation}
\text{Accuracy} = \frac{TP + TN}{TP + FP + TN + FN}
\end{equation}

where $TP$ indicates true positives, $TN$ indicates true negatives, $FP$ indicates false positives, and $FN$ indicates false negatives. Additionally, we make use of the confusion matrix, which offers a tabular overview of accurate and inaccurate predictions for both groups. Additional important measures, namely Sensitivity (True Positive Rate) and Specificity (True Negative Rate), are derived from this:

\begin{equation}
\text{Sensitivity} = \frac{TP}{TP + FN}
\end{equation}

\begin{equation}
\text{Specificity} = \frac{TN}{TN + FP}
\end{equation}

\begin{equation}
\text{F-score} = \frac{2 \cdot TP}{2 \cdot TP + FP + FN}
\end{equation}

We also use the Receiver Operating Characteristic (ROC) curve as a quantitative and visual tool to assess model performance. One important indicator of class separability is the Area Under the Curve (AUC). An AUC of 1.0 indicates complete discrimination between responder and non-responder classes in a perfect model.

\section{Results and Discussion}

\begin{figure}[htbp]
  \centering
  \begin{tabular}{ccc}
    \begin{subfigure}{0.3\linewidth}
      \centering
      \includegraphics[width=\linewidth]{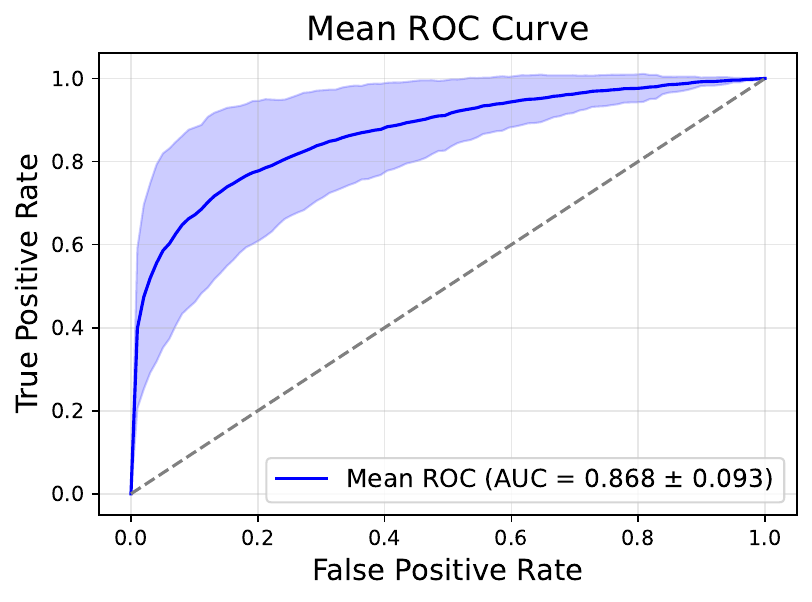}
      \caption{CQT}
    \end{subfigure}
    &
    \begin{subfigure}{0.3\linewidth}
      \centering
      \includegraphics[width=\linewidth]{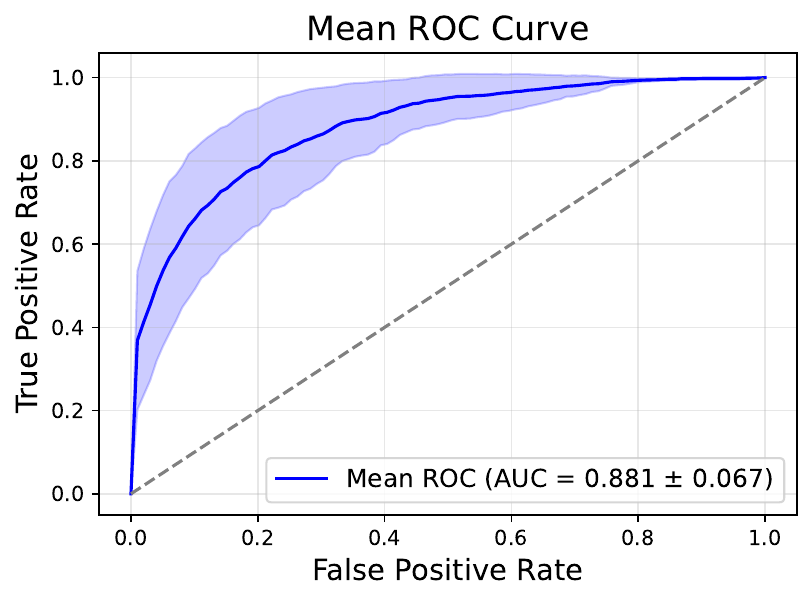}
      \caption{WVD}
    \end{subfigure}
    &
    \begin{subfigure}{0.3\linewidth}
      \centering
      \includegraphics[width=\linewidth]{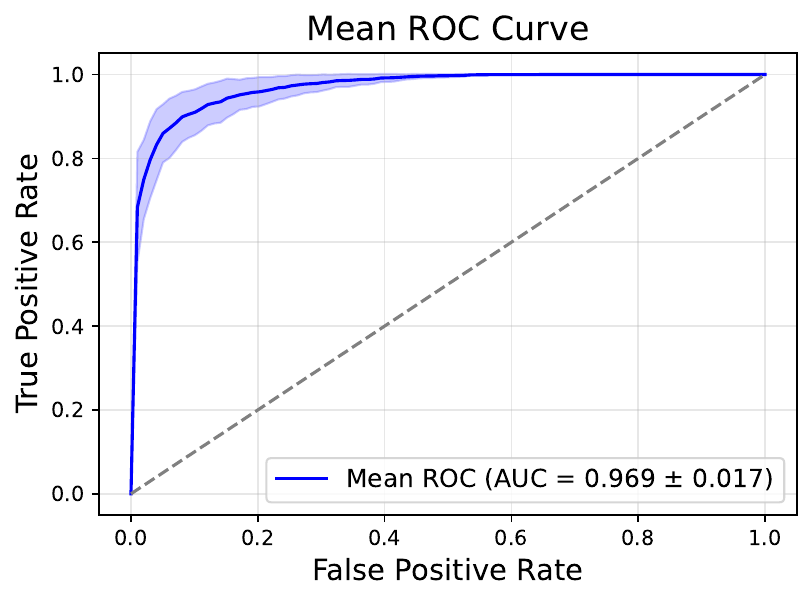}
      \caption{CWT}
    \end{subfigure}
  \end{tabular}
  \caption{Mean ROC curves of three TF image representations with proposed CNN.}
  \label{fig:tf_roc}
\end{figure}

In our first experiment, we evaluated different time–frequency image-based representations to determine which modality best supports our proposed model. All reported performance metrics (AUC, sensitivity, specificity, F1-score, and accuracy) are presented as mean ± standard deviation across the 10-fold cross-validation. As shown in Table~\ref{tab:time_frequency_modalities}, the CWT consistently outperformed the other time-frequency approaches across all evaluation metrics. Specifically, CWT achieved an accuracy of 90.23\% ± 3.2\%, an AUC of 0.9695 ± 0.0178. 
Both CQT and WVD showed very similar performance, but CWT leading in overall stability, highlighting the superior capability of adjust the scale (or frequency) of the wavelet, making it effective for analyzing non-stationary signals like EEG. Fig. \ref{fig:tf_roc} shows the mean ROC curves of CQT, WVD and CWT with the Proposed CNN.

\begin{table}[htbp]
\scriptsize
  \centering
  \caption{Performance comparison of TF image-based representations using the proposed model. CWT achieves the highest overall accuracy.}
  \label{tab:time_frequency_modalities}
  \begin{tabular}{|c|c|c|c|c|c|}
    \hline
    \textbf{Method} &
    \textbf{AUC} &
    \textbf{Sensitivity} & \textbf{Specificity} &
    \textbf{F1-Score} & 
    \textbf{Accuracy}
     \\
    \hline
    CQT & 0.868 ± 0.098 & 0.820 ± 0.100 & 0.761 ± 0.106 & 0.796 ± 0.091 & 0.790 ± 0.092 \\
    \hline
    WVD & 0.881 ± 0.070 & 0.812 ± 0.101 & 0.785 ± 0.090 & 0.800 ± 0.078 & 0.799 ± 0.076  \\
    \hline
    \textbf{CWT} & \textbf{0.969 ± 0.017} & \textbf{0.912 ± 0.053} & \textbf{0.892 ± 0.050} & \textbf{0.903 ± 0.033} & \textbf{0.902 ± 0.032} \\
    \hline
  \end{tabular}
\end{table}

\begin{figure}[htbp]
  \centering
  \begin{tabular}{ccc}
    \begin{subfigure}{0.3\linewidth}
      \centering
      \includegraphics[width=\linewidth]{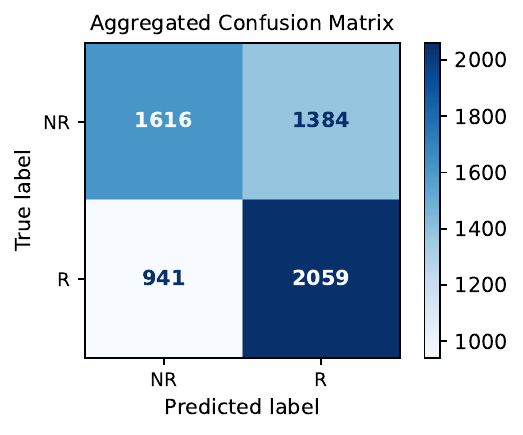}
      \caption{MobileNetV2}
    \end{subfigure}
    &
    \begin{subfigure}{0.3\linewidth}
      \centering
      \includegraphics[width=\linewidth]{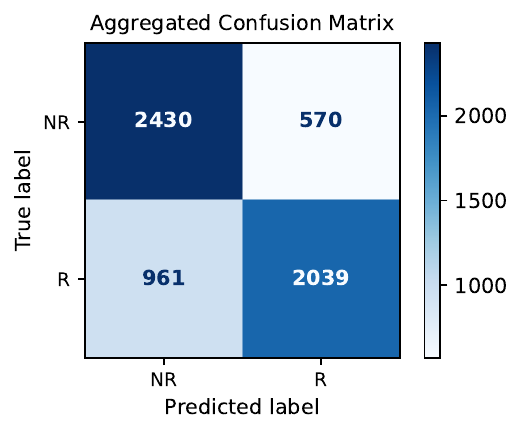}
      \caption{DeepSleepNet}
    \end{subfigure}
    &
    \begin{subfigure}{0.3\linewidth}
      \centering
      \includegraphics[width=\linewidth]{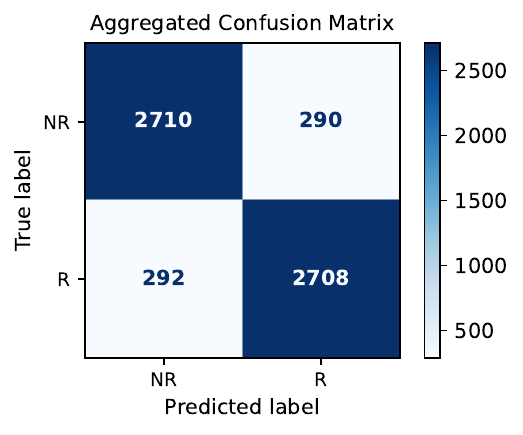}
      \caption{Proposed CNN}
    \end{subfigure}
  \end{tabular}
  \caption{Confusion matrices of two benchmark models and the proposed CNN.}
  \label{fig:confusionmatrix}
\end{figure}

\begin{figure}[htbp]
  \centering
  \begin{tabular}{ccc}
    \begin{subfigure}{0.3\linewidth}
      \centering
      \includegraphics[width=\linewidth]{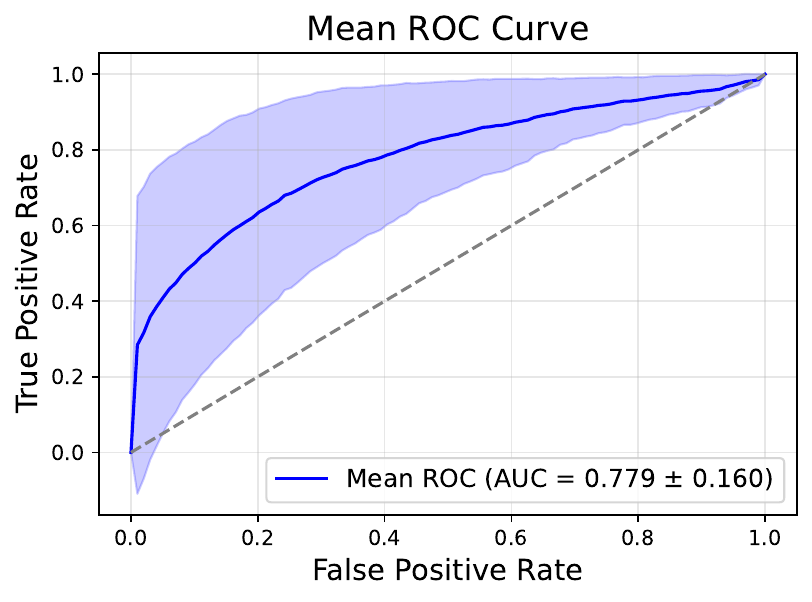}
      \caption{DenseNet201}
    \end{subfigure}
    &
    \begin{subfigure}{0.3\linewidth}
      \centering
      \includegraphics[width=\linewidth]{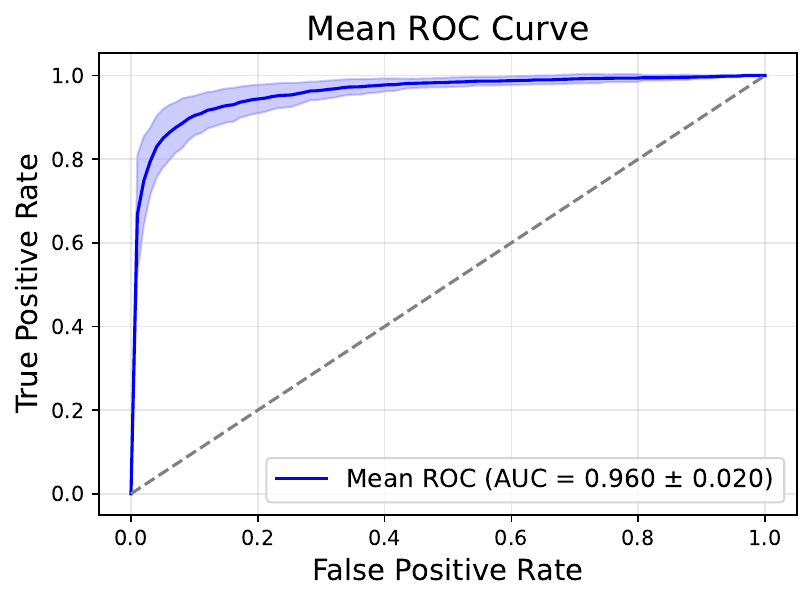}
      \caption{EEGNet}
    \end{subfigure}
    &
    \begin{subfigure}{0.3\linewidth}
      \centering
      \includegraphics[width=\linewidth]{figures/cnn_roc.pdf}
      \caption{Proposed CNN}
    \end{subfigure}
  \end{tabular}
  \caption{Mean ROC curves of two benchmark models and the proposed CNN.}
  \label{fig:roc}
\end{figure}

In the second experiment, we compared the performance of the best-performing time-frequency representation (CWT) against a range of pre-trained and EEG-specific models. As shown in Table~\ref{tab:tf_vs_pretrained}, the proposed CNN consistently outperformed all baselines across every metric, achieving an accuracy of 90.23\% ± 3.2\%, AUC of 0.966 ± 0.022. Among the pre-trained models, DenseNet201 performed best with 71.9\% ± 16.1\% accuracy, followed by MobileNetV2 (61.3\% ± 16.7\%) and Xception (56.9\% ± 20.3\%). Within the EEG-specific group, DeepSleepNet achieved 74.4\% ± 2.1\%, while EEGNet and DeepConvNet reached 89.5\% ± 3.6\% accuracy each. Fig. \ref{fig:confusionmatrix} shows the confusion matrix of MobileNetv2, DeepSleepNet and Proposed CNN. Fig.~\ref{fig:roc} illustrates the mean ROC curves of DenseNet201, EEGNet, and the proposed CNN. These findings highlight the effectiveness of tailored time-frequency features combined with a dedicated CNN architecture, surpassing both general-purpose and domain-specific alternatives.

\begin{table}[htbp]
\scriptsize
  \centering
  \caption{Performance comparison of the best TF representation (CWT) with pretrained and EEG-specific models.}
  \label{tab:tf_vs_pretrained}
  \begin{tabular}{|c|c|c|c|c|c|}
    \hline
    \textbf{Method} &
    \textbf{AUC} &
    \textbf{Sensitivity} & \textbf{Specificity} &
    \textbf{F1-Score} & 
    \textbf{Accuracy}
     \\
    \hline
    Xception & 0.632 ± 0.228 & 0.656 ± 0.184 & 0.482 ± 0.320 & 0.610 ± 0.181 & 0.569 ± 0.203 \\
    \hline
    MobileNetV2 & 0.661 ± 0.203 & 0.686 ± 0.141 & 0.539 ± 0.283 & 0.646 ± 0.147 & 0.613 ± 0.167 \\
    \hline
    DenseNet201 & 0.779 ± 0.169 & 0.809 ± 0.152 & 0.630 ± 0.246 & 0.747 ± 0.150 & 0.719 ± 0.161 \\
    \hline
    DeepSleepNet & 0.795 ± 0.213 & 0.679 ± 0.389 & 0.810 ± 0.191 & 0.660 ± 0.371 & 0.744 ± 0.210 \\
    \hline
    EEGNet & 0.960 ± 0.021 & 0.914 ± 0.035 & 0.876 ± 0.066 & 0.898 ± 0.033 & 0.895 ± 0.036 \\
    \hline
    DeepConvNet & 0.960 ± 0.021 & 0.914 ± 0.035 & 0.876 ± 0.066 & 0.898 ± 0.033 & 0.895 ± 0.036 \\
    \hline
    \textbf{Proposed CNN} & \textbf{0.969 ± 0.017} & \textbf{0.912 ± 0.053} & \textbf{0.892 ± 0.050} & \textbf{0.903 ± 0.033} & \textbf{0.902 ± 0.032} \\
    \hline
  \end{tabular}
\end{table}

Table \ref{tab:comparison_studies} summarizes prior EEG-based approaches for predicting antidepressant response. Earlier studies using handcrafted features or small datasets achieved moderate performance (ACC 87–88\%, SEN 77–95\%, SPE 80–99\%). CWT-based image representations with pretrained ensembles improved results (ACC up to 96.55\%). Our proposed CNN, trained on CWT representations, outperformed all baselines with 90.2\% accuracy, 90\% sensitivity, and 90\% specificity. These results highlight the advantage of deep learning on rich time-frequency EEG features, surpassing both traditional statistical methods and pretrained vision models.

\begin{table}[htbp]
  \centering
  \scriptsize
  \begin{tabular}{|c|p{2.4cm}|p{2.6cm}|c|c|c|c|}
    \hline
    \textbf{Ref.} & \textbf{Methodology} & \textbf{Remarks} & \textbf{Sample Size} & \textbf{ACC (\%)} & \textbf{SEN (\%)} & \textbf{SPE (\%)} \\
    \hline
    \cite{khodayari2013machine} & Nonlinear EEG features, Fisher ratio, MFA classifier & Gender biased [6 female(R), 10 male(NR)] & 7 R / 15 NR & 87.4 & 94.9 & 80.9 \\
    \hline
    \cite{mumtaz2017wavelet} & Time–frequency analysis, ROC, logistic regression & Complex feature space, often difficult for medical practitioners & 16 R / 18 NR & 87.5 & 95 & 80 \\
    \hline
    \cite{jaworska2019leveraging} & Demographics, EEG, source-localized PCA, RF & High Dimensionality and Overfitting Risks & 27 R / 24 NR & 88.0 & 77.0 & 99.0 \\
    \hline
    \cite{korani2026predicting} & FBSE-ED and DWT-based images, a deep learning classifier & No patient-wise splitting, which may affect generalizability & 9 R / 6 NR & 93.60 & 96.33 & 90.85 \\
    \hline
    \cite{shahabi2021prediction} & CWT-based images, ensemble of pretrained models & Lacks subject-wise splitting, which may affect generalizability & 12 R / 18 NR & 96.55 & 96.01 & 96.95 \\
    \hline
    \cite{minami2022association} & LORETA, coherence, functional connectivity & AUC and other metrics are missing & 14 R / 11 NR & – & 82.0 & 86.0 \\
    \hline
    \textbf{Ours} & CWT, CQT, WVD based deep learning & Small sample size, mitigated via cross-validation and subject-wise splitting & 12 R / 18 NR & \textbf{90.3} & \textbf{90.3} & \textbf{90.3} \\
    \hline
  \end{tabular}
  \caption{Comparative analysis of EEG-based SSRI and rTMS response prediction studies.}
  \label{tab:comparison_studies}
\end{table}

This study presents NeuroPredict, a non-invasive EEG-based framework for predicting SSRI treatment response in patients with Major Depressive Disorder. Among the evaluated modalities, CWT achieved the highest performance, with an accuracy of 90.23\% (95\% CI for AUC: 96.3--97.1\%), outperforming both pretrained vision models and EEG-specific baselines. These findings highlight the effectiveness of CWT-based time-frequency representations, which offer stable and noise-resilient feature structures.

Despite the strong performance, the sample size remains relatively modest, which may limit the generalizability of the findings. Future work will focus on validating the framework across larger and more diverse cohorts, as well as incorporating cross-site evaluations. Additionally, integrating multimodal data sources, such as genetic, behavioral, or neuroimaging data, may further enhance robustness and clinical applicability.

%
%
%
%
\bibliographystyle{splncs04}
\bibliography{mybibliography}

@article{zhou2021machine,
  title={Machine learning to predict clinical remission in depressed patients after acute phase selective serotonin reuptake inhibitor treatment},
  author={Zhou, Shuzhe and Ma, Qinhong and Lou, Yiwei and Lv, Xiaozhen and Tian, Hongjun and Wei, Jing and Zhang, Kerang and Zhu, Gang and Chen, Qiaoling and Si, Tianmei and others},
  journal={Journal of Affective Disorders},
  volume={287},
  pages={372--379},
  year={2021},
  publisher={Elsevier}
}

@article{li2025neurophysiological,
  title={Neurophysiological mechanisms and predictive modeling of SSRI treatment response in depression disorder based on multidimensional EEG features},
  author={Li, Gang and Huang, Boyi and Wang, Yuling and Zhou, Bin and Hu, Fo and Wang, Linbing},
  journal={Journal of Affective Disorders},
  pages={120424},
  year={2025},
  publisher={Elsevier}
}

@article{yadav2025evaluating,
  title={Evaluating the impact of machine learning models on adult major depressive disorder using conventional treatment strategies: a systematic review approach},
  author={Yadav, Nishant and Gulati, Anamika and Gulati, Varun and Yadav, Prashant},
  journal={Discover Public Health},
  volume={22},
  number={1},
  pages={410},
  year={2025},
  publisher={Springer}
}

@inproceedings{korani2026predicting,
  title={Predicting the Outcome of rTMS Depression Therapy Using EEG Signals and CNN},
  author={Korani, Wael and Chowdhury, Md Fahimul Kabir and AlQadi, Sadam and Kumar, Priyan Malarvizhi and Rostami, Reza and Kazemi, Reza},
  booktitle={Recent Trends in Image Processing and Pattern Recognition},
  year={2026},
  publisher={Springer},
  }

@article{albizu2024machine,
  title={Machine learning-optimized non-invasive brain stimulation and treatment response classification for major depression},
  author={Albizu, Alejandro and Indahlastari, Aprinda and Suen, Paulo and Huang, Ziqian and Waner, Jori L and Stolte, Skylar E and Fang, Ruogu and Brunoni, Andre R and Woods, Adam J},
  journal={Bioelectronic Medicine},
  volume={10},
  number={1},
  pages={25},
  year={2024},
  publisher={Springer}
}

@article{korani2026fusion,
  title={Fusion Techniques of Time Frequency-based Images to Predict the Outcome of rTMS Depression Therapy},
  author={Korani, Wael and Chowdhury, Md Fahimul Kabir and Aledhari, Mohammed and Rostami, Reza and Kazemi, Reza},
  journal={Biomedical Signal Processing and Control},
  year={2026},
  publisher={Elsevier}
}

@inproceedings{chowdhury2026multifrequency,
  title={Multi-Frequency EEG Fusion and Phase-Based Connectivity for Predicting rTMS Outcomes in Depression},
  author={Chowdhury, Md Fahimul Kabir and Korani, Wael and Rostami, Reza and Kazemi, Reza},
  booktitle={International Conference on Neural Information Processing},
  year={2026},
  organization={Springer}
}

@article{lawhern2018eegnet,
  title={EEGNet: a compact convolutional neural network for EEG-based brain--computer interfaces},
  author={Lawhern, Vernon J and Solon, Amelia J and Waytowich, Nicholas R and Gordon, Stephen M and Hung, Chou P and Lance, Brent J},
  journal={Journal of neural engineering},
  volume={15},
  number={5},
  pages={056013},
  year={2018},
  publisher={iOP Publishing}
}

@article{schirrmeister2017deep,
  title={Deep learning with convolutional neural networks for EEG decoding and visualization},
  author={Schirrmeister, Robin Tibor and Springenberg, Jost Tobias and Fiederer, Lukas Dominique Josef and Glasstetter, Martin and Eggensperger, Katharina and Tangermann, Michael and Hutter, Frank and Burgard, Wolfram and Ball, Tonio},
  journal={Human brain mapping},
  volume={38},
  number={11},
  pages={5391--5420},
  year={2017},
  publisher={Wiley Online Library}
}

@article{lin2025integrating,
  title={Integrating multi-polygenic scores for enhanced prediction of antidepressant treatment outcomes in an East Asian population},
  author={Lin, Shu-Chin and Fang, Chiu-Ping and Hsu, Chia-Lin and Chung, An-Nie and Chen, Tzu-Ting and Hsu, Kai-Hsiang and Yeh, Chueh-Chun and Zheng, Jingyi and Liu, ChaoYu and Wu, Chi-Shin and others},
  journal={Neuropsychopharmacology},
  pages={1--8},
  year={2025},
  publisher={Springer International Publishing Cham}
}

@inproceedings{chowdhury2025brain,
  title={Brain tumor classification in mri images: A computationally efficient convolutional neural network},
  author={Chowdhury, Md Fahimul Kabir and Ferdous, Jannatul},
  booktitle={2025 IEEE International Conference on Biomedical Engineering, Computer and Information Technology for Health (BECITHCON)},
  pages={633--638},
  year={2025},
  organization={IEEE}
}

@article{byun2025automated,
  title={Automated classification of stress and relaxation responses in major depressive disorder, panic disorder, and healthy participants via heart rate variability},
  author={Byun, Sangwon and Kim, Ah Young and Shin, Min-Sup and Jeon, Hong Jin and Cho, Chul-Hyun},
  journal={Frontiers in Psychiatry},
  volume={15},
  pages={1500310},
  year={2025},
  publisher={Frontiers Media SA}
}

@article{wang2025predicting,
  title={Predicting the Treatment Response of Patients With Major Depressive Disorder to Selective Serotonin Reuptake Inhibitors Using Machine Learning Techniques and EEG Functional Connectivity Features},
  author={Wang, Fanglan and You, Zifan and Zhang, Tingkai and Xu, Kai and Wang, Liangliang and He, Jingqi and Tang, Jinsong},
  journal={Depression and Anxiety},
  volume={2025},
  number={1},
  pages={9340993},
  year={2025},
  publisher={Wiley Online Library}
}

@inproceedings{moncy2024prediction,
  title={Prediction of Treatment Outcome to Transcranial Direct Current Stimulation in Major Depression Based on Deep Learning of EEG Data},
  author={Moncy, Jijomon Chettuthara and Fan, Yong and Fu, Cynthia HY},
  booktitle={2024 IEEE Conference on Artificial Intelligence (CAI)},
  pages={1123--1128},
  year={2024},
  organization={IEEE}
}

@article{shahabi2025deep,
  title={Deep Vision Transformer Model for the Prediction of Antidepressants Outcome in Major Depressive Disorder Using Raw EEG Signals.},
  author={Shahabi, Mohsen Sadat and Shalbaf, Ahmad},
  journal={Frontiers in Biomedical Technologies},
  volume={12},
  year={2025}
}

@article{mumtaz2017wavelet,
  title={A wavelet-based technique to predict treatment outcome for major depressive disorder},
  author={Mumtaz, Wajid and Xia, Likun and Mohd Yasin, Mohd Azhar and Azhar Ali, Syed Saad and Malik, Aamir Saeed},
  journal={PloS one},
  volume={12},
  number={2},
  pages={e0171409},
  year={2017},
  publisher={Public Library of Science San Francisco, CA USA}
}

@article{khodayari2013machine,
  title={A machine learning approach using EEG data to predict response to SSRI treatment for major depressive disorder},
  author={Khodayari-Rostamabad, Ahmad and Reilly, James P and Hasey, Gary M and De Bruin, Hubert and MacCrimmon, Duncan J},
  journal={Clinical Neurophysiology},
  volume={124},
  number={10},
  pages={1975--1985},
  year={2013},
  publisher={Elsevier}
}

@article{jaworska2019leveraging,
  title={Leveraging machine learning approaches for predicting antidepressant treatment response using electroencephalography (EEG) and clinical data},
  author={Jaworska, Natalia and De la Salle, Sara and Ibrahim, Mohamed-Hamza and Blier, Pierre and Knott, Verner},
  journal={Frontiers in psychiatry},
  volume={9},
  pages={768},
  year={2019},
  publisher={Frontiers Media SA}
}

@article{shahabi2021prediction,
  title={Prediction of drug response in major depressive disorder using ensemble of transfer learning with convolutional neural network based on EEG},
  author={Shahabi, Mohsen Sadat and Shalbaf, Ahmad and Maghsoudi, Arash},
  journal={Biocybernetics and Biomedical Engineering},
  volume={41},
  number={3},
  pages={946--959},
  year={2021},
  publisher={Elsevier}
}

@article{minami2022association,
  title={Association between the rostral anterior cingulate cortex and anterior insula in the salience network on response to antidepressants in major depressive disorder as revealed by isolated effective coherence},
  author={Minami, Shota and Kato, Masaki and Ikeda, Shunichiro and Yoshimura, Masafumi and Ueda, Satsuki and Koshikawa, Yosuke and Takekita, Yoshiteru and Kinoshita, Toshihiko and Nishida, Keiichiro},
  journal={Neuropsychobiology},
  volume={81},
  number={6},
  pages={475--483},
  year={2022},
  publisher={S. Karger AG}
}

@book{stahl2021stahl,
  title={Stahl's essential psychopharmacology: neuroscientific basis and practical applications},
  author={Stahl, Stephen M},
  year={2021},
  publisher={Cambridge university press}
}

@article{cipriani2018comparative,
  title={Comparative efficacy and acceptability of 21 antidepressant drugs for the acute treatment of adults with major depressive disorder: a systematic review and network meta-analysis},
  author={Cipriani, Andrea and Furukawa, Toshi A and Salanti, Georgia and Chaimani, Anna and Atkinson, Lauren Z and Ogawa, Yusuke and Leucht, Stefan and Ruhe, Henricus G and Turner, Erick H and Higgins, Julian PT and others},
  journal={The Lancet},
  volume={391},
  number={10128},
  pages={1357--1366},
  year={2018},
  publisher={Elsevier}
}

@article{supratak2017deepsleepnet,
  title={DeepSleepNet: A model for automatic sleep stage scoring based on raw single-channel EEG},
  author={Supratak, Akara and Dong, Hao and Wu, Chao and Guo, Yike},
  journal={IEEE transactions on neural systems and rehabilitation engineering},
  volume={25},
  number={11},
  pages={1998--2008},
  year={2017},
  publisher={IEEE}
}

@article{Mumtaz2016,
author = "Wajid  Mumtaz",
title = "{MDD Patients and Healthy Controls EEG Data (New)}",
year = "2016",
month = "11",
url = "https://figshare.com/articles/dataset/EEG_Data_New/4244171",
doi = "10.6084/m9.figshare.4244171.v2"
}
\end{document}